\documentclass[10pt,twocolumn,letterpaper]{article}

\usepackage{cvpr}
\usepackage{times}
\usepackage{graphicx} 
\usepackage{amsmath}
\usepackage{amssymb}

\usepackage{booktabs}
\usepackage{multirow}
\makeatletter
\newcommand{\figcaption}[1]{\def\@captype{figure}\caption{#1}}
\newcommand{\tblcaption}[1]{\def\@captype{table}\caption{#1}}
\makeatother

\usepackage[pagebackref=true,breaklinks=true,letterpaper=true,colorlinks,bookmarks=false]{hyperref}

\cvprfinalcopy 

\def\cvprPaperID{3905} 

\ifcvprfinal\fi
\begin{document}

\title{Between-class Learning for Image Classification}

\author{\\
Institution1\\
Institution1 address\\
{\tt\small firstauthor@i1.org}
\and
Second Author\\
Institution2\\
First line of institution2 address\\
{\tt\small secondauthor@i2.org}
}

\author{Yuji Tokozume$^1$\,\,\,\,\,\,\,\,\,\,\,\,Yoshitaka Ushiku$^1$\,\,\,\,\,\,\,\,\,\,\,\,Tatsuya Harada$^1$$^,$$^2$\\
$^1$The University of Tokyo\,\,\,\,\,\,\,\,\,\,\,\,$^2$RIKEN\\
{\tt\small \{tokozume,ushiku,harada\}@mi.t.u-tokyo.ac.jp}\\
}

\maketitle
\thispagestyle{empty}

\begin{abstract}
In this paper, we propose a novel learning method for image classification called Between-Class learning (BC learning)\footnote{Similar idea {\it mixup} \cite{zhang2017mixup} was proposed on Oct.~25, 2017 (unpublished), but our preliminary experimental results on CIFAR-10 and ImageNet-1K were already presented in ILSVRC2017 on July 26, 2017.}. 
We generate between-class images by mixing two images belonging to different classes with a random ratio. We then input the mixed image to the model and train the model to output the mixing ratio.
BC learning has the ability to impose constraints on the shape of the feature distributions, and thus the generalization ability is improved.
BC learning is originally a method developed for sounds, which can be digitally mixed.
Mixing two image data does not appear to make sense; however, we argue that because convolutional neural networks have an aspect of treating input data as waveforms, what works on sounds must also work on images.
First, we propose a simple mixing method using internal divisions, which surprisingly proves to significantly improve performance. Second, we propose a mixing method that treats the images as waveforms, which leads to a further improvement in performance. As a result, we achieved $19.4\%$ and $2.26\%$ top-1 errors on ImageNet-1K and CIFAR-10, respectively.\footnote{The code is publicly available at \newline \url{https://github.com/mil-tokyo/bc_learning_image/}.}
\end{abstract}


\section{Introduction}
Deep convolutional neural networks (CNNs) \cite{lecun1998gradient} have achieved high performance in various tasks, such as image recognition \cite{krizhevsky2012imagenet, he2016deep}, speech recognition \cite{abdel2014convolutional, sainath2015learning}, and sound recognition \cite{piczak2015environmental, tokozume2017learning}. One of the biggest themes of research on image recognition has been network engineering. Many types of image recognition networks have been proposed mainly in ILSVRC competition \cite{krizhevsky2012imagenet, simonyan2015very, szegedy2015going, he2016deep, xie2017aggregated, zhang2016polynet, hu2017squeeze}. Furthermore, training deep neural networks is difficult, and many techniques have been proposed to achieve a high performance: data augmentation techniques \cite{krizhevsky2012imagenet}, new network layers such as dropout \cite{srivastava2014dropout} and batch normalization \cite{ioffe2015batch}, optimizers such as Adam \cite{kingma2014adam}, and so on. Thanks to these research studies, training deep neural networks has become relatively easy, with a stable performance, at least for image classification. At present, a novel approach is needed for further improvement.

\begin{figure}
\begin{center}
	\includegraphics[width=0.9\hsize, clip]{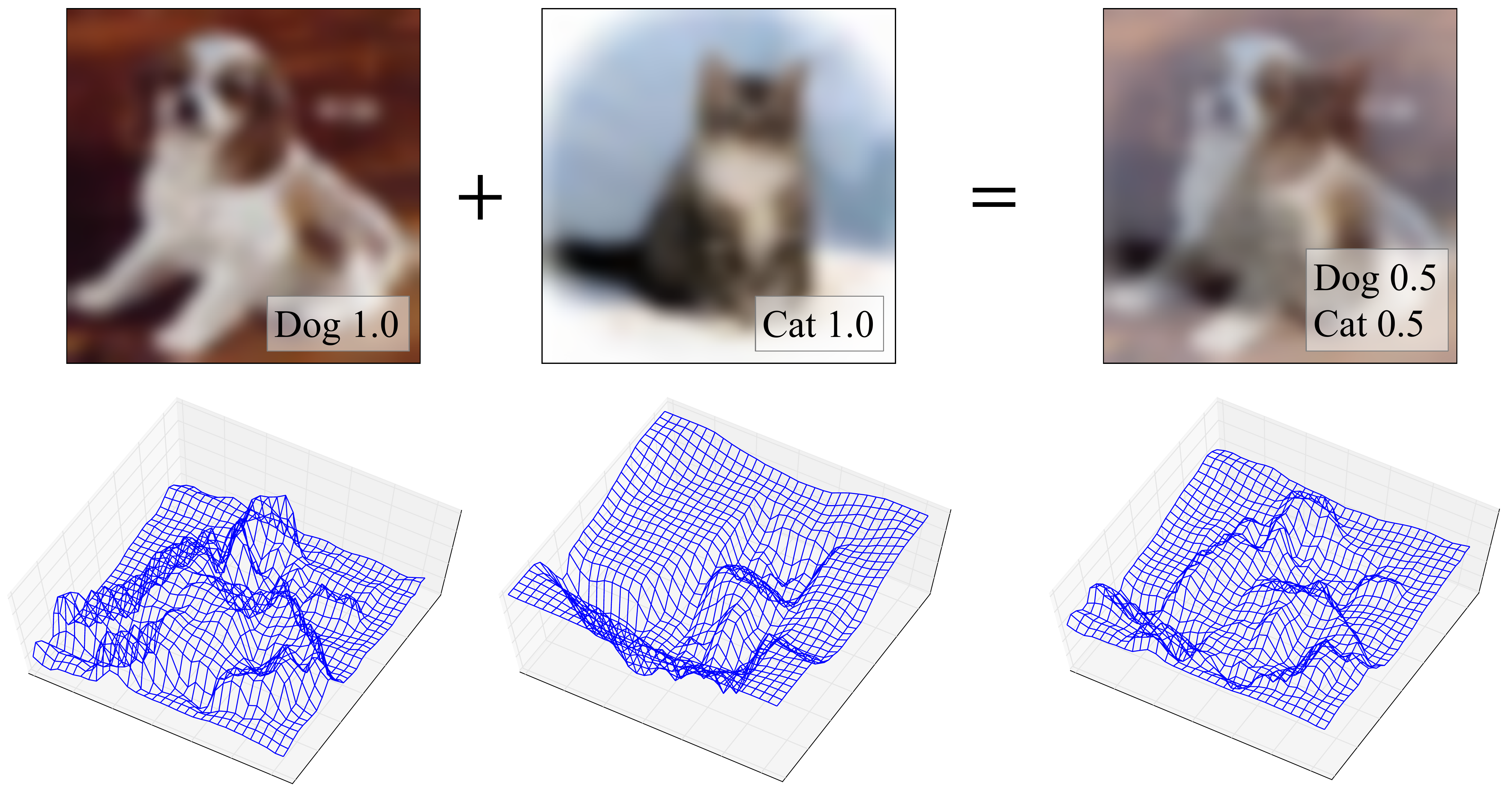}
\end{center}
\vspace{-3mm}
   \caption{We argue that CNNs have an aspect of treating the input data as waveforms. In that case, a mixture of two images is a mixture of two waveforms. This make sense {\it for machines}, although it does not visually make sense {\it for humans}.}
\label{fig:wave}
\vspace{-4mm}
\end{figure}

In \cite{tokozume2018learning}, a simple and powerful learning method named {\it Between-Class learning (BC learning)} was proposed for deep sound recognition. BC learning aims to learn a classification problem by solving the problem of predicting the mixing ratio between two different classes. They generated between-class examples by mixing two sounds belonging to different classes with a random ratio. They then input the mixed sound to the model and trained the model to output the mixing ratio of each class. The advantages of BC learning are not limited only to the increase in variation of the training data. They argued that BC learning has the ability to impose constraints on the feature distributions, which cannot be achieved with standard learning, and thus the generalization ability is improved. They carefully designed the method of mixing two sounds, considering the difference in the sound energies, to achieve a satisfactory performance. As a result, BC learning improved the performance on various sound recognition networks, datasets, and data augmentation schemes, and they achieved a performance surpasses the human level in sound classification tasks.

The question here is whether BC learning also performs well on images. The core idea of BC learning itself, i.e., mixing two examples and training the model to output the mixing ratio, can be used irrespective of the modality of input data. BC learning is applicable to sounds, because sound is a kind of wave motion and a mixture of multiple sound data still counts as a sound. However, an image is not a kind of wave motion, and mixing multiple image data does not visually make sense. We show an example of a mixed image in Fig.~\ref{fig:wave}(top). A mixed image loses its objectness and does not count as an image. Therefore, it appears inappropriate to apply BC learning to images.

However, the important thing is not how {\it humans} perceive the mixed data, but how {\it machines} perceive them. We argue that CNNs have an aspect of treating input data as waveforms, considering the recent studies on speech and sound recognition and the characteristics of image data as pixel values. We assume that CNNs recognize images by treating them as waveforms in quite a similar manner to how they recognize sounds. Thus, a mixture of two images is a mixture of two waveforms as shown in Fig.~\ref{fig:wave}(bottom), and it would make sense {\it for machines}. Therefore, what is effective for sounds would also be effective for images.

We thus propose BC learning for images in this paper. First, we propose the simplest mixing method using internal divisions. Surprisingly, this mixing method proves to perform well. Second, we also propose an improved mixing method that treats images as waveform data (BC+). In this method, we subtract the per-image mean value from each image. By doing this, we can treat each image as a zero-mean waveform similar to a sound. We then define the image energy as the standard deviation per image, and mix two images considering the image energy in quite a similar manner to what \cite{tokozume2018learning} did for sounds. This mixing method is also simple and easy to implement, and leads to a further improvement in performance.

Experimental results show that BC learning also improves the performance of various types of image recognition networks from a simple network to the state-of-the-art networks. The top-$1$ error of ResNeXt-101 ($64\times4$d) \cite{xie2017aggregated} on ImageNet-1K is improved from $20.4\%$ to $19.4\%$ by using the simplest BC learning. Moreover, the error rate of the state-of-the-art Shake-Shake Regularization \cite{gastaldi2017shake} on CIFAR-10 dataset is improved from $2.86\%$ to $2.26\%$ by using the improved BC learning (BC+). Finally, we visualize the learned features and show that BC learning indeed imposes a constraint on the feature distribution. The contributions of this paper are as follows:
\begin{itemize}
\vspace{-1mm}
\setlength{\itemsep}{0cm} 
 \item{We applied BC learning \cite{tokozume2018learning} to images by mixing two images belonging to different classes and training the model to output the mixing ratio.}
 \item{We argued that CNNs have an aspect of treating input data as waveforms, and proposed a mixing method that treats the images as waveforms.}
 \item{We conducted experiments extensively and demonstrated the effectiveness of BC learning for images.}
\end{itemize}

The remainder of this paper is organized as follows. In Section \ref{2}, we provide a summary of BC learning for sound recognition \cite{tokozume2018learning} as a related work. We then propose BC learning for image recognition in Section \ref{3}, explaining the relationship with BC learning for sounds. In Section \ref{4}, we compare the performance of standard learning and BC learning, and demonstrate the effectiveness of BC learning. Finally, we conclude this paper in Section \ref{5}.


\section{BC learning for sounds}\label{2}
In this section, we describe BC learning for sound recognition \cite{tokozume2018learning} as a related work. The contents in this section is a summarization of Section 3 of \cite{tokozume2018learning}. Please see \cite{tokozume2018learning} for more detailed information.  

\subsection{Overview}
In standard learning for classification problems, a single training example is selected from the dataset and input to the model. Then, the the model is trained to output a one-hot label. By contrast, in BC learning, two training examples belonging to different classes are selected from the dataset and mixed with a random ratio. Then, the mixed data is input to the model, and the model is trained to output the mixing ratio of each class. KL-divergence between the outputs of the model and the ratio labels is used as the loss function, instead of the usual cross-entropy loss. Note that mixing is not performed in testing phase.


\subsection{Mixing method}\label{mixsound}
Let $\{{\bf x}_1,\, {\bf t}_1\}$ and $\{{\bf x}_2,\, {\bf t}_2\}$ be two sets of sounds and one-hot labels belonging to different classes randomly selected from the training dataset. A random ratio $\, r \,$ is generated from the uniform distribution $\, U(0,\ 1) \,$, and two sounds and labels are mixed with this ratio. Two labels $\, {\bf t}_1\, $ and $\, {\bf t}_2 \,$ are mixed simply by $\, r \, {\bf t}_1 + (1-r) \, {\bf t}_2 \,$ because BC learning aims to train the model to output the mixing ratio. We then explain how to mix two sounds $\, {\bf x}_1 \,$ and $\, {\bf x}_2 \,$, which should be carefully designed to achieve a satisfactory performance. The simplest method is $\, r \, {\bf x}_1 + (1-r) \, {\bf x}_2 \,$. Here, $\frac{r \, {\bf x}_1 + (1-r) \, {\bf x}_2}{\sqrt{r^2 + (1-r)^2}}\,$ is better because sound energy is proportional to the square of the amplitude. However, when the difference in sound pressure level between $\, {\bf x}_1 \,$ and $\, {\bf x}_2 \,$ is large, the perception of the sound mixed by this method does not become $\, {\bf x}_1:{\bf x}_2=r:(1-r) \,$. In this case, training the model with a label of $\, r \, {\bf t}_1 + (1-r) \, {\bf t}_2 \,$ is inappropriate. To address this problem, they proposed to use a mixing method that considers the sound pressure level of two sounds $\, G_1 \,$ and $\, G_2 \,$ [dB] so that the auditory perception of the mixed sound becomes $\, {\bf x}_1:{\bf x}_2=r:(1-r) \,$:

\begin{equation}
\label{eqn:mixsound}
\begin{split}
  \frac{p \, {\bf x}_1 + (1-p) \, {\bf x}_2}{\sqrt{p^2 + (1-p)^2}}, \hspace{10mm} \\ 
  {\rm where} \;\; p = \frac{1}{1\,+\, 10^{\frac{G_1-G_2}{20}} \cdot \frac{1\,-\,r}{r}}.
\end{split}
\end{equation}

\subsection{How BC learning works}
We explain how BC learning improves the classification performance. They argued that BC learning has the ability to impose the following two constraints on the feature distributions learned by the model by training the model to output the mixing ratio between two classes:
\begin{itemize}
\vspace{-1mm}
\setlength{\itemsep}{0cm} 
 \item{Enlargement of Fisher's criterion \cite{fisher1936use} ({\it i.e.}, the ratio of the between-class distance to the within-class variance).}
 \item{Regularization of positional relationship among feature distributions.}
\end{itemize}

They hypothesized that when a mixed sound is input to the model, the feature of the mixed sound is located approximately in near the internally dividing point of the features of original two sounds. This hypothesis came from the fact that linearly-separable features are learned in hidden layers close to the output layer \cite{an2015can} and that humans can perceive which of the two sounds is louder from a digitally mixed sound. They showed that this hypothesis was indeed correct by visualizing the feature distributions of the standard-learned model using PCA. Under this hypothesis, we explain the two constraints that BC learning can impose on the feature distributions.

\begin{figure}
	\centering
	\includegraphics[width=0.95\hsize, clip]{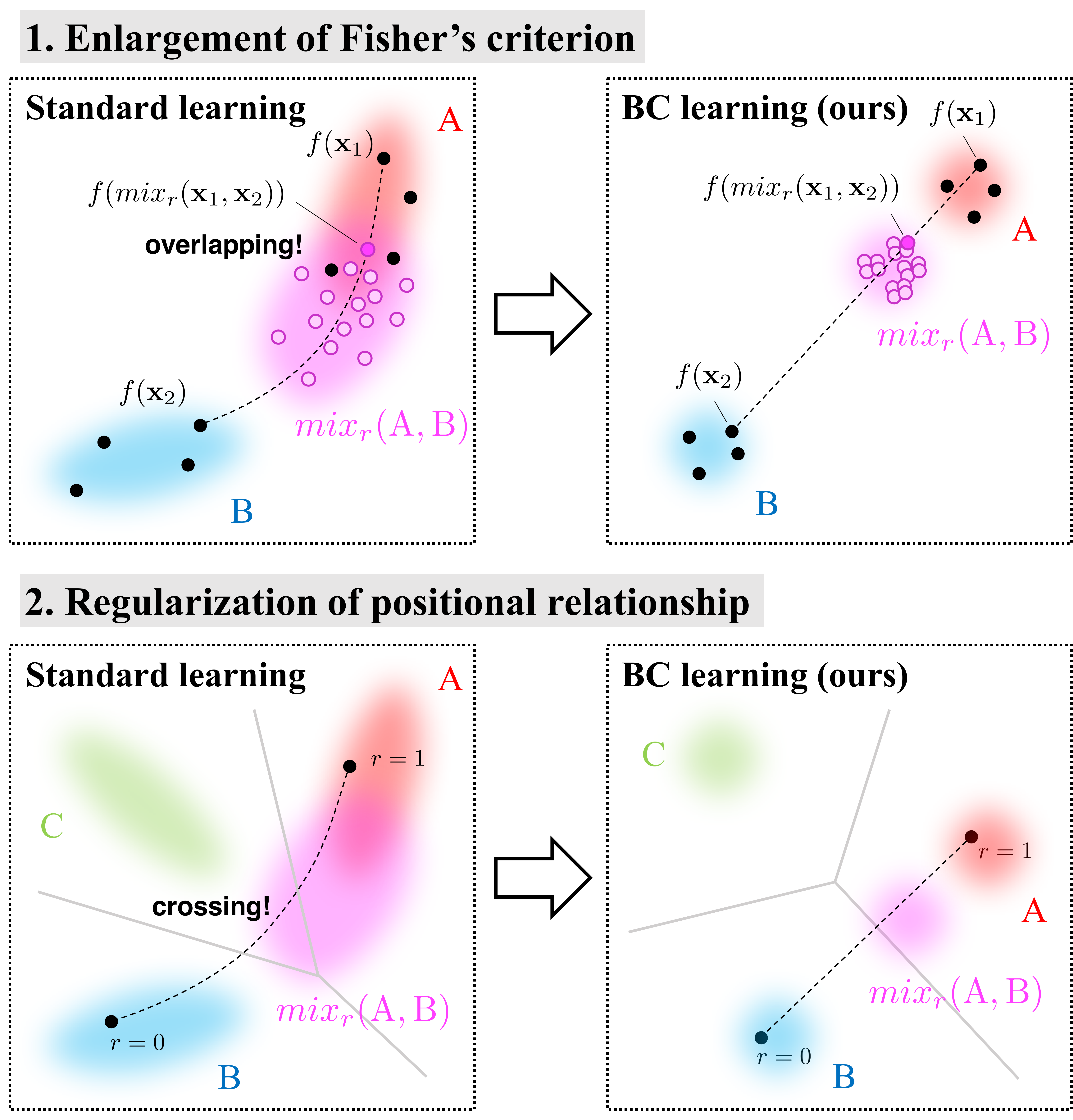}
	\vspace{2mm}
	\caption{BC learning has the ability to impose constraints on the feature distribution, which cannot be achieved with the standard learning \cite{tokozume2018learning}. This figure represents the class distribution in the feature space. The black dashed line represents the trajectory of the feature when we input a mixture of two particular sounds to the model changing the mixing ratio from $0$ to $1$.}
	\label{fig:bcsounds}
	\vspace{-2mm}
\end{figure}

\paragraph{Enlargement of Fisher's criterion.}\label{fisher}
They argued that BC learning enlarges Fisher's criterion \cite{fisher1936use} between any two classes in the feature space. We explain the reason in Fig.~\ref{fig:bcsounds}(top). If Fisher's criterion between the feature distributions of class A (red) and class B (blue) is small as shown in Fig.~\ref{fig:bcsounds}(upper left), the feature distribution of the sounds obtained by mixing class A and B at a certain ratio (magenta) becomes large, and would have a large overlap with one or both feature distributions of class A and class B. In this case, the model cannot output the mixing ratio for some mixed examples projected onto the overlapping area, and BC learning gives a large loss. To let the model output the mixing ratio and make the loss of BC learning small, Fisher's criterion should be large as shown in Fig.~\ref{fig:bcsounds}(upper right). In this case, the overlap becomes small, and BC learning gives a small loss. Therefore, BC learning enlarges Fisher's criterion in the feature space.

\paragraph{Regularization of positional relationship among feature distributions.}
They also argued that BC learning has the effect of regularizing the positional relationship among class feature distributions. We explain the reason in Fig.~\ref{fig:bcsounds}(bottom). If the features of each class are not regularly distributed as shown in Fig.~\ref{fig:bcsounds}(lower left), the decision boundary of class C other than both A and B would appear between class A and class B, and some of the mixed sounds of class A and class B would be misclassified into class C. This is an undesirable situation because there is little possibility that a mixed sound of two classes becomes a sound of other classes. BC learning gives a large loss to this situation because BC learning trains the model to output the mixing ratio between class A and class B. If the features of each class are regularly distributed as shown in Fig.~\ref{fig:bcsounds}(lower right), on the other hand, the decision boundary of class C does not appear between class A and class B, and the model can output the mixing ratio instead of misclassifying the mixed sound as class C. As a result, the loss of BC learning becomes small. Therefore, BC learning has the effect of regularizing the positional relationship of the feature distributions. In this way, they argued that BC learning has the ability to impose constraints on the feature distribution, and thus BC learning improves the generalization ability.

\section{From sounds to images}\label{3}
In this section, we consider applying BC learning to images. Following BC learning for sounds \cite{tokozume2018learning}, we select two training examples from different classes and mix these two examples using a random ratio. We then input the mixed data to the model and train the model to output the mixing ratio. BC learning uses only mixed data and labels and, thus, never uses pure data and labels. How to mix two examples is also important for images. First, we propose the simplest mixing method in Section \ref{v1}. Second, we discuss why BC learning can also be applied to images in Section \ref{why}. Finally, in Section \ref{v2}, we propose a better version of BC learning, considering the discussion in \ref{why}.

\subsection{Simple mixing}\label{v1}
Let $\, {\bf x}_1\, $ and $\, {\bf x}_2 \,$ be two images belonging to different classes randomly selected from the training dataset, and $\, {\bf t}_1\, $ and $\, {\bf t}_2 \,$ be their one-hot labels. Note that $\, {\bf x}_1\, $ and $\, {\bf x}_2 \,$ may have already been preprocessed, and they have the same size as the input size of the network. We generate a random ratio $\, r \,$ from $\, U(0,\ 1) \,$, and mix two sets of data and labels with this ratio. Because we aim to train the model to output the mixing ratio, we mix two labels simply by:
\begin{equation}
r \, {\bf t}_1 + (1-r) \, {\bf t}_2.
\end{equation}

We now explain how to mix $\, {\bf x}_1 \,$ and $\, {\bf x}_2 \,$. In \cite{tokozume2018learning}, a carefully designed mixing method of two sounds was proposed considering the difference in the sound energies, as we mentioned in Section \ref{mixsound}. In a sound data, $0$ is the absolute center, and the distance from $0$ represents the sound energy. However, the pixel values of an image data do not have an absolute center, and there appears to be no concept of energy. We thus first propose the following mixing method as the simplest method using internal divisions\footnote{{\it mixup} \cite{zhang2017mixup} used this mixing method.}:
\begin{equation}
 \label{eqn:v1}
  r \, {\bf x}_1 + (1-r) \, {\bf x}_2.
\end{equation}

\subsection{Why BC learning works on images}\label{why}
BC learning is applicable to sounds because a mixed sound still counts as a sound. Sound is a kind of wave motion, and mixing two sounds physically makes sense. Humans can recognize two sounds and perceive which of the two sounds is louder or softer from the digitally mixed sound. However, image data, as pixel values, is not a kind of wave motion {\it for humans}. Therefore, mixing multiple images does not visually make sense. 

However, the important thing is not whether mixing two data physically makes sense, or whether humans can perceive a mixed data, but how a machine perceives a mixed data. We argue that CNNs have an aspect of treating input data as waveforms. In fact, recent studies have demonstrated that CNNs can learn speech and sounds directly from raw waveforms, and each filter learns to respond to a particular frequency area \cite{sainath2015learning, tokozume2017learning, dai2017very}. It is also known that images, as pixel values, can be transformed into components of various frequency areas by using $2$-D Fourier transform \cite{shanmugam1979optimal}, and some convolutional filters can act as frequency filters ({\it e.g.}, a Laplacian filter acts as a high-pass filter \cite{burt1983laplacian}). Therefore, it is expected that each convolutional filter of a CNN learns to extract the frequency features. In this way, we assume that CNNs recognize images by treating them as waveforms in quite a similar manner to how they recognize sounds. Thus, because a mixture of two images is a mixture of two waveforms {\it for machines}, what is effective for sounds would also be effective for images.

\begin{figure}
	\centering
	\includegraphics[width=\hsize]{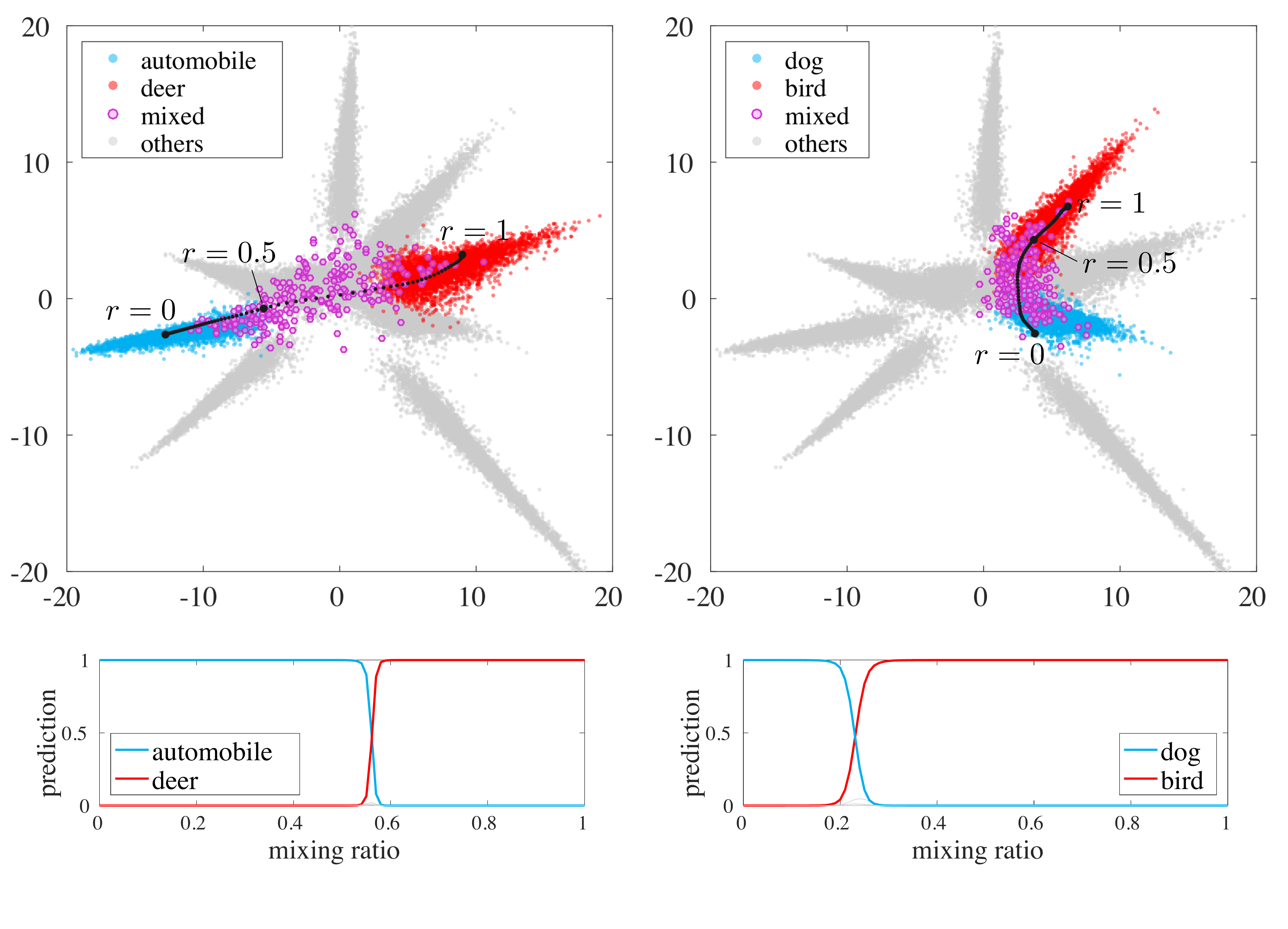}
	\vspace{-7mm}
	\caption{Visualization of the feature space of a standard-learned model using PCA. The features of the mixed images are distributed between two classes.}
	\label{fig:pca}
	\vspace{-2mm}
\end{figure}

We visualized the feature distribution of the standard-learned model against mixed data using PCA. We used the output of the $10$-th layer of an 11-layer CNN trained on CIFAR-10. We mixed two images using Eqn.~\ref{eqn:v1} and the results are shown in Fig.~\ref{fig:pca}. The magenta dots represent the feature distribution of the mixed images of {\it automobile} and {\it deer} (left), {\it dog} and {\it bird} (right) with a ratio of $0.5:0.5$, and the black dotted line represents the trajectory of the feature when we input a mixture of two particular images to the model, changing the mixing ratio from $0$ to $1$. This figure shows that the mixture of two images is projected onto the point near the internally dividing point of two features, and the features of the mixed images are distributed between two classes, which is the same tendency observed for sounds \cite{tokozume2018learning}. Therefore, the same effect as BC learning for sounds, i.e., an enlargement of Fisher's criterion in the feature space and a regularization of the positional relationship among the feature distributions of the classes, is expected. We compare the feature distributions learned with standard and BC learning and demonstrate that a different shape of feature distribution is learned with BC learning in the visualization experiment.

\subsection{BC+: Images as waveform data}\label{v2}
Here, we consider a new mixing method, which treats images as waveform data. We regard image data as a $2$-D waveform consisting of (R, G, B) vectors. In recent state-of-the-art methods, the input data is normalized for each channel using the mean and standard deviation calculated from the whole training data \cite{xie2017aggregated, huang2017densely, gastaldi2017shake}. In this case, the mean of each image is not equal to $0$, and each image data ${\bf x}_i$ is represented as ${\bf x}_i = \mu_i + {\bf d}_i$, where $\, \mu_i \,$ and $\, {\bf d}_i \,$ are the static component and wave component, respectively. Here, the simplest mixing method of Eqn.~\ref{eqn:v1} can be rewritten as $\{r \, \mu_1 + (1-r) \, \mu_2\} +   \{r \, {\bf d}_1 + (1-r) \, {\bf d}_2\}$. We assume that the performance improvement with Eqn.~\ref{eqn:v1} is mainly owing to the wave component $\, r \, {\bf d}_1 + (1-r) \, {\bf d}_2$ if CNNs treat the input data as waveforms. Moreover, the static component $\,r \, \mu_1 + (1-r) \, \mu_2$ can have a bad effect because mixing two waveforms generally hypothesizes that the static components of two waveforms are same. 

Therefore, we remove the static component by subtracting the per-image mean value (not channel-wise mean). We consider the following mixing method instead of Eqn.~\ref{eqn:v1}. By doing this, we can treat each image as a zero-mean waveform similar to a sound.
\begin{equation}
 \label{eqn:v1.0}
  r \, ({\bf x}_1 - \mu_1) + (1-r) \, ({\bf x}_2 - \mu_2)
\end{equation}

We then apply a scheme similar to that applied to sounds as described in Section \ref{mixsound}. First, we consider mixing two images with:
\begin{equation}
 \label{eqn:v1.5}
  \frac{r \, ({\bf x}_1-\mu_1) + (1-r) \, ({\bf x}_2-\mu_2)}{\sqrt{r^2 + (1-r)^2}},
\end{equation}
instead of Eqn.~\ref{eqn:v1.0}, considering that waveform energy is proportional to the square of the amplitude. This process prevents the input variance from decreasing.

Second, we take the difference of image energies into consideration, in order to make the perception of the mixed image $\, {\bf x}_1:{\bf x}_2=r:(1-r) \,$. We use a new coefficient $\,p \,$ instead of $\, r \,$ and mix two images by $\frac{p \, ({\bf x}_1 - \mu_1) + (1-p) \, ({\bf x}_2 - \mu_2)}{\sqrt{p^2 + (1-p)^2}}\,$. We define $\, p \,$ using the standard deviation per image ($\sigma_1$ and $\sigma_2$) so that the ratio of amplitude becomes $\, {\bf x}_1:{\bf x}_2=r:(1-r) \,$. We solve $\, p \, \sigma_1 : (1 - p)  \, \sigma_2 = r:(1-r) \,$ and obtain the proposed mixing method:
\begin{equation}
\begin{split}
 \label{eqn:v2}
  \frac{p \, ({\bf x}_1-\mu_1) + (1-p) \, ({\bf x}_2-\mu_2)}{\sqrt{p^2 + (1-p)^2}}, \\ 
  {\rm where} \;\; p = \frac{1}{1\,+\, \frac{\sigma_1}{\sigma_2} \cdot \frac{1\,-\,r}{r}}. \hspace{4mm}
\end{split}
\end{equation}

The main differences from the mixing method for sounds (Eqn.~\ref{eqn:mixsound}) are subtracting per-image mean values and using standard deviations instead of sound pressure levels. This mixing method is also easy to implement, and experimentally proves to lead to a further improvement in performance, compared to the simplest mixing method of Eqn.~\ref{eqn:v1}.


\section{Experiments}\label{4}
\subsection{Experiments on ImageNet-1K}
\begin{table}
	\centering
	\caption{Results of ResNeXt-101 ($64\times4$d) \cite{xie2017aggregated} on ImageNet-1K dataset. BC learning improves the performance when using the default learning schedule. Furthermore, the performance is further improved when using a longer learning schedule, and the single-crop  top-1 error is improved by around $1\%$ compared to the default performance reported in \cite{xie2017aggregated}.}
	\label{tab:imagenet}
	\vspace{2mm}
	\small
	\begin{tabular}{llcc}
		\toprule
		&& \multicolumn{2}{c}{Top-1/top-5 val. error ($\%$)} \\
		\cmidrule{3-4}
		\# epochs & Learning & Single-crop & 10-crop \\
		\midrule
		\multirow{2}{*}{$100$} & Standard & $20.4$ / $5.3$ \cite{xie2017aggregated} & $18.90$ / $4.61$ \\	
						& BC (ours) & $19.92$ / $4.91$ & $18.66$ / $4.26$ \\
		\midrule
		\multirow{2}{*}{$150$} & Standard & $20.44$ / $5.25$ & $18.98$ / $4.43$\\
						& BC (ours) & ${\bf 19.43}$ / ${\bf 4.80}$ & ${\bf 18.22}$ / ${\bf 4.13} $\\
		\bottomrule
	\end{tabular}
\end{table} 

\begin{figure}
	\centering
	\vspace{-4mm}
	\includegraphics[width=0.95\hsize, clip]{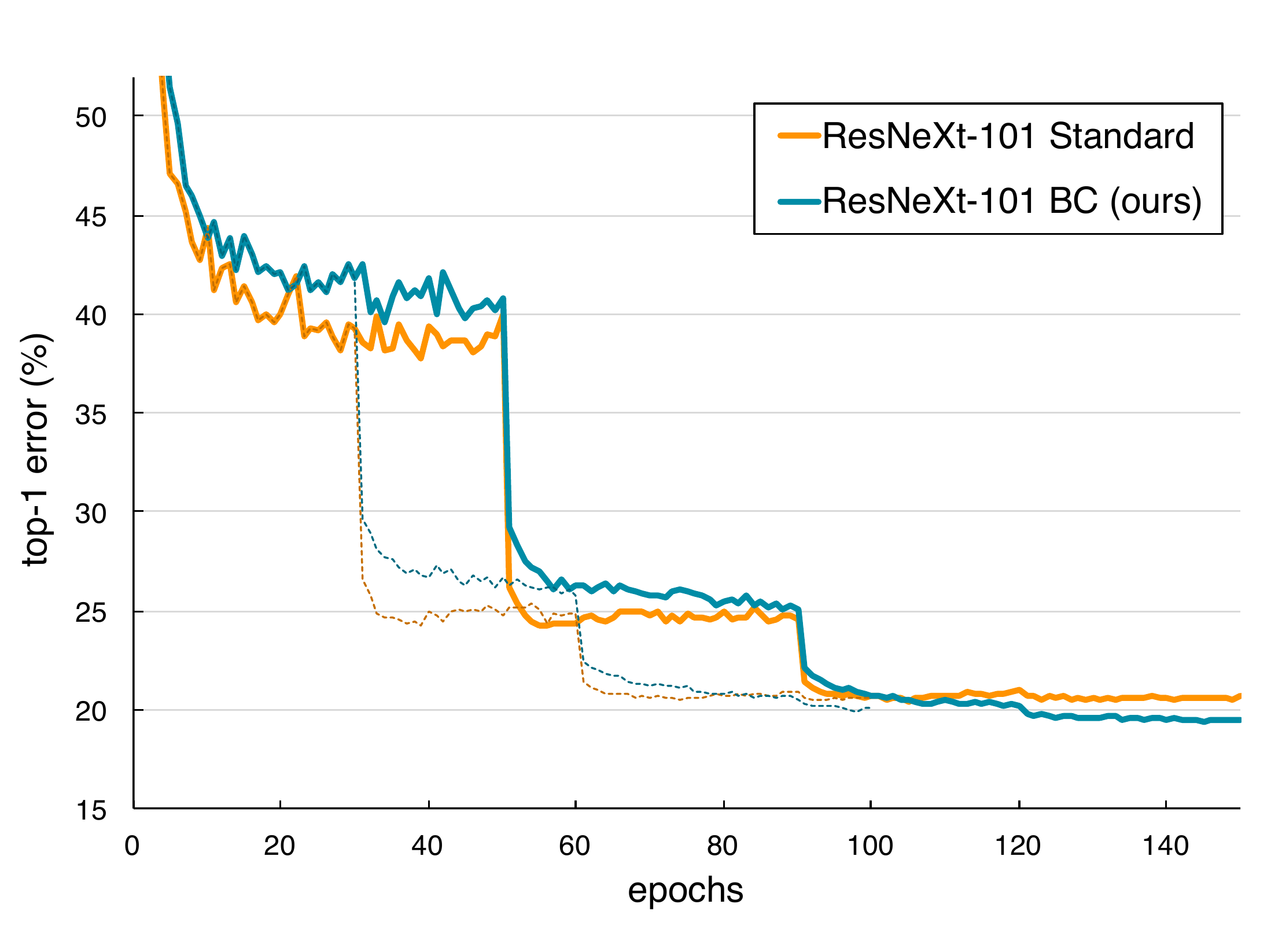}	
		\caption{Training curves of ResNeXt-101 ($64\times4$d) on ImageNet-1K dataset. The dashed lines represent the training curves when using the default learning schedule, and the solid lines represent the training curves when using a longer learning schedule.}
		\label{fig:imagenet}
	\vspace{-2mm}
\end{figure}

\begin{table*}
	\centering
	\caption{Results on CIFAR-10 and CIFAR-100 datasets. We show the average and the standard error of $5$ or $10$ trials. BC learning improves the performance of various settings. Note that $^{\dagger}$ is trained with a different learning setting from the default.}
	\label{tab:cifar}
	\vspace{2mm}
	\small
	\begin{tabular}{llcc}
		\toprule
		&& \multicolumn{2}{c}{Error rate ($\%$) on} \\
		\cmidrule{3-4}
		Model & Learning & CIFAR-10 & CIFAR-100 \\
		\midrule
		\multirow{3}{*}{$11$-layer CNN} & Standard & $6.07 \pm0.04 $ & $26.68 \pm0.09$ \\
						 		   & BC (ours) & $5.40 \pm0.07$ & $24.28 \pm0.11$ \\
								   & BC+ (ours) & ${\bf 5.22 \pm0.04}$ & ${\bf 23.68 \pm0.10}$ \\
		\midrule						   
								   
		\multirow{3}{*}{ResNet-29$^\dagger$ \cite{xie2017aggregated}} & Standard & $4.24 \pm 0.06$ / $4.39$ \cite{xie2017aggregated} & $20.18 \pm0.07$ \\
						 		   & BC (ours) & $3.75 \pm0.04$ & $19.56 \pm0.10$ \\
								   & BC+ (ours) & ${\bf 3.55 \pm0.03}$ & ${\bf 19.41 \pm0.07}$ \\
		\midrule
		\multirow{3}{*}{ResNeXt-29 ($16\times64$d)$^\dagger$ \cite{xie2017aggregated}} & Standard & $3.54 \pm 0.04$ / $3.58$ \cite{xie2017aggregated} & ${\bf 16.99 \pm0.06} $ / $17.31$ \cite{xie2017aggregated} \\
						 		   & BC (ours) & ${\bf 2.79 \pm0.06}$ & $18.21 \pm0.12$ \\
								   & BC+ (ours) & $2.81 \pm0.06$ & $17.93 \pm0.09$ \\
		\midrule
		\multirow{3}{*}{DenseNet-BC ($k=40$)$^\dagger$ \cite{huang2017densely}} & Standard & $3.61 \pm 0.10$ / $3.46$ \cite{huang2017densely} & $17.28 \pm0.12 $ / $17.18$ \cite{huang2017densely} \\
						 		   & BC (ours) & $2.68 \pm0.03$ & $16.36 \pm0.10$ \\
								   & BC+ (ours) & ${\bf 2.57 \pm0.06}$ & ${\bf 16.23 \pm0.07}$ \\
		\midrule
		\multirow{3}{*}{Shake-Shake Regularization \cite{gastaldi2017shake}} & Standard & $2.86$ \cite{gastaldi2017shake} & ${\bf 15.85}$ \cite{gastaldi2017shake} \\
						 		   & BC (ours) & $2.38 \pm0.04$ & $15.90 \pm0.06$  \\
								   & BC+ (ours) & ${\bf 2.26 \pm0.01}$ & $16.00 \pm0.10$ \\
		\bottomrule
	\end{tabular}
	\vspace{-2mm}
\end{table*} 

First, we compare the performance of standard and BC learning on the $1{,}000$-class ImageNet classification task \cite{russakovsky2015imagenet}. In this experiment, we used the simple BC learning. We selected ResNeXt-101 ($64\times4$d) \cite{xie2017aggregated} as the model for training because it has a state-of-the-art level performance and, moreover, the official Torch \cite{collobert2002torch} training codes are available\footnote{https://github.com/facebookresearch/ResNeXt}. To validate the comparison, we incorporated BC learning into these official codes. When using BC learning, we selected two training images and applied the default data augmentation scheme as in described \cite{xie2017aggregated} to each image, and obtained two $224\times224$ images. We then mixed these two images with a random ratio selected from $U(0, \, 1)$. In addition to the default learning schedule (\# of epochs = $100$), we also tried a longer learning schedule (\# of epochs = $150$). In $100$-epochs training, we started training with a learning rate of $0.1$ and then divided the learning rate by $10$ at the epoch in $\{30,\, 60,\, 90\}$, as in \cite{xie2017aggregated}. In $150$-epochs training, we also started training with a learning rate of $0.1$ and then divided the learning rate by $10$ at the epoch in $\{50,\, 90,\, 120,\, 140\}$. We reported classification errors on the validation set using both single-crop testing \cite{xie2017aggregated} and 10-crop testing \cite{krizhevsky2012imagenet}.

The results are shown in Table \ref{tab:imagenet}. We also show the training curves in Fig.~\ref{fig:imagenet}. The performance of BC learning with the default $100$-epochs training was significantly improved from that of standard learning. Moreover, the performance of BC learning was further improved with $150$-epochs training, while that of standard learning was not improved, and we achieved $19.43\%$/$4.80\%$ single-crop top-1/top-5 validation errors and $18.22\%$/$4.13\%$ $10$-crop validation top1/top5 errors. The single-crop top-1 error was improved by around $1\%$ compared to the default performance reported in \cite{xie2017aggregated}.

\paragraph{Discussion.}
Learning between-class examples among $1{,}000$ classes is difficult, and it tends to require a large number of training epochs. As shown in Fig.~\ref{fig:imagenet}, the performance on the first $100$ epochs of the $150$-epochs training of BC learning is worse than the performance of standard learning. Therefore, the learning schedule should be carefully designed. Furthermore, we assume that the usage of curriculum learning \cite{bengio2009curriculum} would be helpful to speed up the training; namely, at the early stage, we generate a mixing ratio close to $0$ or $1$ and input relatively pure examples, and we gradually change the distribution of $\, r\,$ to flat.

\subsection{Experiments on CIFAR}
\begin{figure*}
\begin{center}
	\includegraphics[width=\hsize, clip]{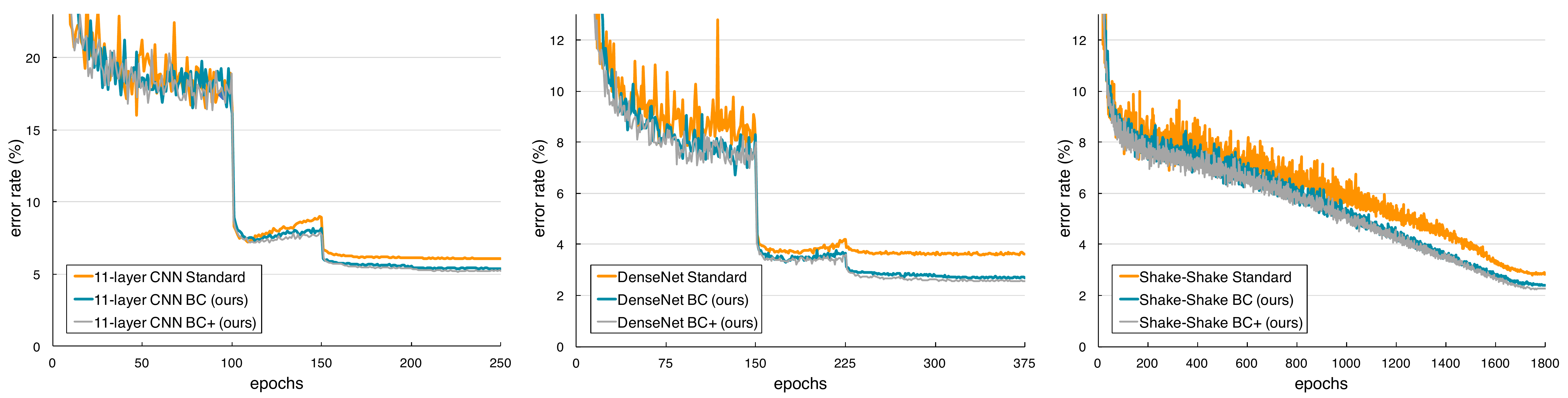}
\end{center}
\vspace{-3mm}
   \caption{Training curves on CIFAR-10 (average of all trials).}
\label{fig:cifar}
\vspace{-2mm}
\end{figure*}

Now, we compare the performance of standard and BC learning on CIFAR-10 and CIFAR-100 datasets \cite{krizhevsky2009learning}. We trained the standard 11-layer CNN, ResNet-29 \cite{xie2017aggregated}, ResNeXt-29 ($16\times64$d) \cite{xie2017aggregated}, DenseNet (BC, $k=40$) \cite{huang2017densely}, and Shake-Shake Regularization (S-S-I and S-E-I for CIFAR-10 and CIFAR-100, respectively) \cite{gastaldi2017shake}. To validate the comparison, we also incorporated BC learning into their original Torch \cite{collobert2002torch} training codes\footnote{https://github.com/facebookresearch/ResNeXt, https://github.com/\newline liuzhuang13/DenseNet, https://github.com/xgastaldi/shake-shake}. The 11-layer CNN was also incorporated into one of them. All of these codes use the standard shifting/mirroring data augmentation scheme that is widely used for these two datasets. We added $75$ training epochs with a further smaller learning rate (1/10) to the default learning schedule of a total of $300$ epochs for ResNet-29, ResNeXt-29, and DenseNet. We show the difference from the default settings in the appendix, as well as the configuration of 11-layer CNN. We trained each model $10$ times for the 11-layer CNN and ResNet-29, and $5$ times for other networks. We report the average and the standard error of the final top-1 errors.

We summarize the results in Table \ref{tab:cifar}. The performances of all networks on CIFAR-10 were improved with the simple BC learning. Furthermore, with the improved version of BC learning (BC+), which treat the image data as waveforms, the performance was further improved. The best result on CIFAR-10 was $2.26\%$ on Shake-Shake Regularization. The performance was stable, and the error rate of all $5$ trials were in the range of $2.25\%$--$2.28\%$. We do not know whether this result should be counted as the state-of-the-art; however, BC learning proves to be able to improve the performance of various networks, from a simple network to the state-of-the-art networks.

We also show the training curves in Fig.~\ref{fig:cifar}. Note that the training curves represent the average of all trials. Contrary to the training curves on ImageNet-1K, the testing error of BC learning decreases at almost the same speed as the standard learning in the early stage of the training. Furthermore, the last $75$ training epochs for 11-layer CNN and DenseNet leads to a lower testing error when using BC learning.

The performances on CIFAR-100 were also improved with BC learning. Although it may be difficult to learn the between-class examples among $100$ classes with no improvement to performance on ResNeXt-29 and Shake-Shake Regularization, BC learning shows a significant improvement on 11-layer CNN, ResNet-29, and DenseNet.

\paragraph{Relationship with data augmentation.}
\begin{table}[b]
	\centering
	\vspace{-3mm}
	\caption{Comparison when using no data augmentation. The figures between brackets indicate the error rates when using the standard data augmentation.}
	\label{tab:augment}
	\vspace{2mm}
	\small
	\begin{tabular}{llcc}
		\toprule
		&& \multicolumn{2}{c}{Error rate ($\%$) on} \\
		Model & Learning & CIFAR-10 & CIFAR-100 \\
		\midrule
		\multirow{2}{*}{11-layer CNN} & Standard	& $9.68$ ($6.07$) & $33.04$ ($26.68$) \\
						 		& BC	 (ours)	& $8.38$ ($5.40$) & $31.00$ ($24.28$) \\
		\midrule
		\multirow{2}{*}{ResNet-29 \cite{xie2017aggregated}} & Standard	& $8.38$ ($4.24$) & $31.36$ ($20.18$) \\
						 						& BC (ours)	& $7.69$ ($3.75$) & $30.79$ ($19.56$)\\
		\bottomrule
	\end{tabular}
\end{table}

Here, we show the performance when using no data augmentation in Table \ref{tab:augment}. We show the average of $10$ trials. As shown in this table, the degree of improvement in the performance is at the same level as, or even smaller than when using the standard data augmentation, although the variation of training data increases from $50{,}000$ to approximately $_{50{,}000}C_2$. We assume this is because the potential within-class variance is small when using no data augmentation. If the within-class variance of the feature space is small, the variance of the features of the mixed images also becomes small, and the overlap in Fig.~\ref{fig:bcsounds}(top) becomes small. Therefore, the effect of BC learning becomes small as a result. We assume that BC learning is compatible with, or even strengthened by, a strong data augmentation scheme.

\subsection{Ablation analysis}
To understand the part that is important for BC learning, we conducted an ablation analysis following \cite{tokozume2018learning}. We trained an 11-layer CNN on CIFAR-10 and CIFAR-100 using various settings. We implemented the codes of ablation analysis using Chainer v1.24 \cite{tokui2015chainer}. All results are shown in Table \ref{tab:ablation}, as well as the average of $10$ trials.

\paragraph{Mixing method.}
The differences of the improved BC learning (Eqn.~\ref{eqn:v2}) from the simplest BC learning (Eqn.~\ref{eqn:v1}) are as follows: a) we subtract par-image mean; b) we divide the mixed image by $\sqrt{r^2 + (1-r)^2}$ considering that waveform energy is proportional to the square of the amplitude; and c) we take the difference of image energies into consideration. We investigated which of them has a great effect. As a result, considering the difference of image energies (c) proved to be of little significance, comparing a+b and a+b+c. This would be because the variance of image energies is smaller than that of sound energies. However, per-image mean subtraction (a) and dividing by the square root (b) are important (a+b+c {\it vs.} b+c and a {\it vs.} a+b). This result shows that treating the image data as waveforms contributes to the improvement in performance.

\paragraph{Label.}
We compared the different labels that we applied to the mixed image. The performance worsened when we used a single label (${\bf t}={\bf t}_1$ if $r>0.5$, otherwise ${\bf t}={\bf t}_2$) and softmax cross entropy loss. It would be inappropriate to train the model to recognize a mixed image as particular class. Using multi label (${\bf t}={\bf t}_1\,+\,{\bf t}_2$) and sigmoid cross entropy loss marginally improved the performance, but the proposed ratio label and KL loss performed the best. The model can learn the between-class examples more efficiently when using our ratio label.

\begin{table}
	\centering
	\caption{Ablation analysis. We trained 11-layer CNN on CIFAR-10 and CIFAR-100 using various settings. We report the average error rate of $10$ trials.}
	\label{tab:ablation}
	\vspace{2mm}
	\small
	\begin{tabular}{llcc}
		\toprule
		&& \multicolumn{2}{c}{Error rate ($\%$) on} \\
		Comparison of & Setting & C-10 & C-100 \\
		\midrule
		\multirow{5}{*}{Mixing method}
					& None (Eqn.~\ref{eqn:v1}, BC)    & $5.40$ & $24.28 $ \\
					& a (Eqn.~\ref{eqn:v1.0})          & $5.45$ & $24.25 $ \\
					& a+b (Eqn.~\ref{eqn:v1.5})       & ${\bf 5.17} $ &$23.72$ \\
					& a+b+c (Eqn.~\ref{eqn:v2}, BC+)   & $5.22$ & ${\bf 23.68}$ \\
					& b+c          & $5.26$ &$23.98$ \\
		\midrule
		\multirow{3}{*}{Label} & Single	 		& $6.35$ & $27.28$ \\
						 & Multi			& $6.05$ & $26.31$ \\
			 			& Ratio (proposed)	& ${\bf 5.22}$ & ${\bf 23.68}$ \\
		\midrule
		\multirow{5}{*}{\# mixed classes} & $N=1$ 		      & $5.98$ & $26.01$ \\
								     & $N=1$ or $2$ 	      & $5.31$ & $23.79$ \\
	 							     & $N=2$ (proposed)  & $5.22$ & ${\bf 23.68}$ \\
								     & $N=2$ or $3$ 	     & ${\bf 5.15}$ & $23.78$ \\
								     & $N=3$  		      & $5.32$ & $24.20$ \\
		\midrule
		\multirow{6}{*}{Where to mix} 	& Input (proposed)   & ${\bf 5.40}$ & $24.28$ \\
								& pool1 		        & $5.74$ & ${\bf 24.09} $ \\
								& pool2 			& $6.52$ & $25.38 $ \\
								& pool3 			& $6.05$ & $27.40 $ \\
								& fc4 			& $6.05$ & $26.70 $ \\
								& fc5				& $6.12$ & $25.99 $ \\
		\midrule
		\midrule
		\multicolumn{2}{c}{Standard learning} & $6.07$ & $26.68$ \\
		\bottomrule
	\end{tabular}
	\vspace{-3mm}
\end{table}

\paragraph{Number of mixed classes.}
We investigated the relationship between the performance and the number of classes of images that we mixed. Surprisingly, the performance was improved when we mixed two images belonging to the same class ($N=1$). Additionally, if we selected two images completely randomly and allowed the two images to be sometimes the same class ($N=1$ or $2$), the performance was worse than proposed $N=2$, in which two images were always different classes. Because BC learning is a method of providing constraints to the feature distribution of different classes, we should select two images belonging to the different classes. We also tried to use mixtures of three different classes with a probability of $0.5$ in addition to the mixtures of two different classes ($N=2$ or $3$), but the performance was not significantly improved from $N=2$. Moreover, the performance when we used only the mixtures of three different classes ($N=3$) was worse than that of $N=2$ despite the larger variation in training data. We assume that mixing more than two classes cannot efficiently provide constraints to the feature distribution.

\paragraph{Where to mix.}
We investigated what occurs when we mix two examples within the network. Here, we used the simple mixing method of Eqn.~\ref{eqn:v1}. The performance was also improved when we mixed two examples at the layer near the input layer. We assume this is because the activations of lower layers can be treated as waveforms because the spatial information is preserved to some extent. Additionally, mixing at the layer close to the output layer had little effect on the performance. It is interesting that the performance was worsened when we mixed two examples at the middle point of the network (pool2 and pool3 for CIFAR-10 and CIFAR-100, respectively). It is expected that the middle layer of the network extracts features that represent both spatial and semantic information simultaneously, and mixing such features would not make sense for machines.

\subsection{Visualization}
\begin{figure}
	\centering
	\vspace{-3mm}
	\includegraphics[width=0.98\hsize, clip]{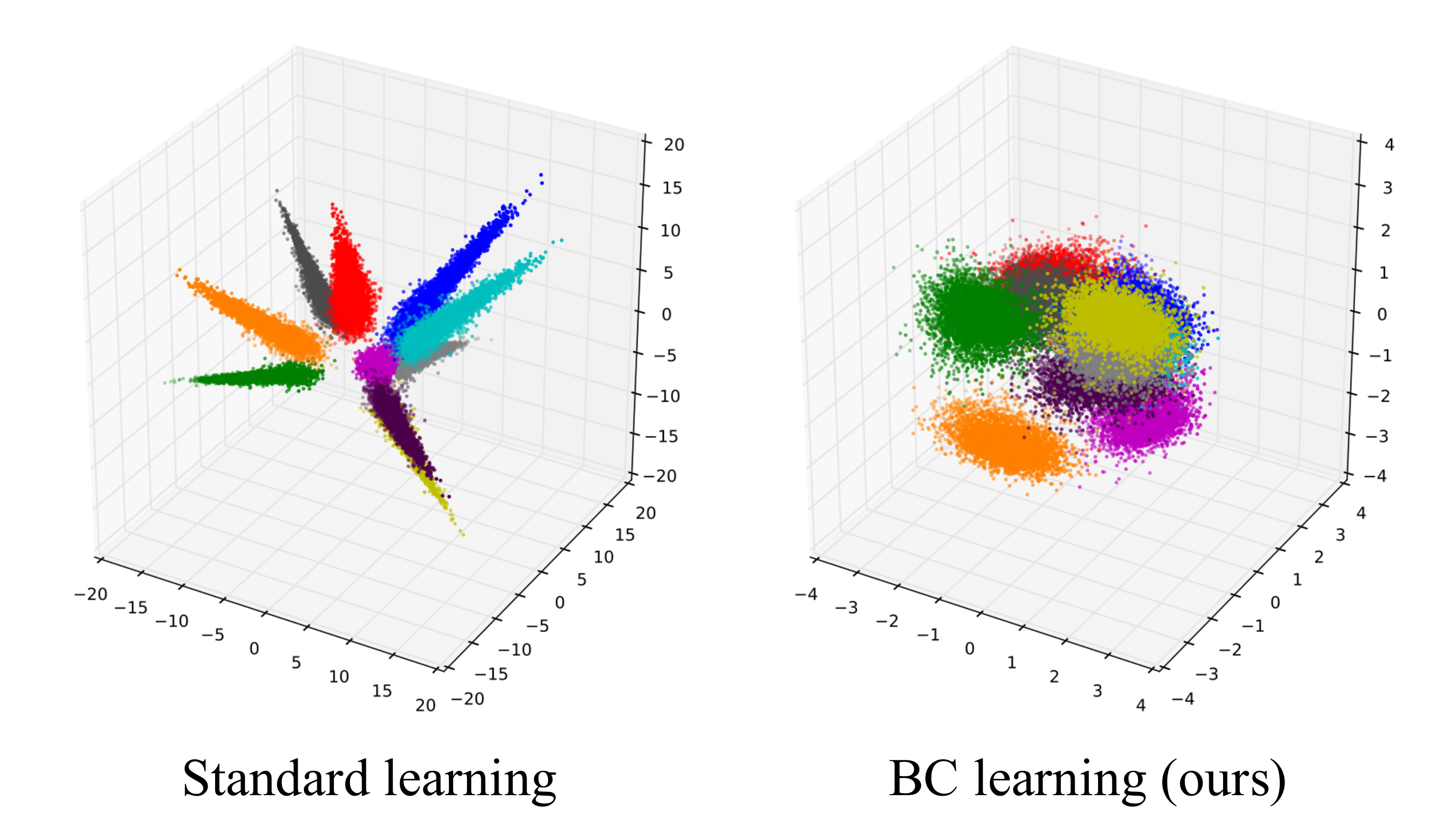}
	\vspace{-1mm}
	\caption{Visualization of feature distributions using $3$-D PCA. BC learning indeed imposes a constraint on the feature distribution.}
	\label{fig:feature}
	\vspace{-3mm}
\end{figure}

Finally, we visualize the features learned with standard and BC learning in Fig.~\ref{fig:feature}. We applied PCA to the activations of the $10$-th layer of the 11-layer CNN trained on CIFAR-10 against the training data. As shown in this figure, the features obtained with BC learning are spherically distributed, and have small within-class variances, whereas that obtained with standard learning are widely distributed from near to far the decision boundaries. We conducted further analysis on the learned features in the appendix. In this way, BC learning indeed imposes a constraint on the feature distribution, which cannot be achieved with standard learning. We conjecture that is why the classification performance was improved with BC learning.


\section{Conclusion}\label{5}
We proposed a novel learning method for image classification called BC learning. We argued that CNNs have an aspect of treating input data as waveforms, and attempted to apply a similar idea to what has been done for sounds. As a result, the performance was significantly improved by simply mixing two images using internal divisions and training the model to output the mixing ratio. Moreover, the performance was further improved with a mixing method that treats the images as waveforms. BC learning is a simple and powerful method that can impose constraints on the feature distribution. We assume that BC learning can be applied not only to images and sounds but also to other modalities.

\section*{Acknowledgement}
This work was supported by JST CREST Grant Number JPMJCR1403, Japan.

\newpage

{\small
\bibliographystyle{ieee}
\bibliography{egpaper_for_review}
}


\newpage
\appendix

\section{Analysis on learned features}
In Fig.~6 of the main paper, we visualized the learned features of 11-layer CNN and demonstrated that BC learning has the ability to impose a constraint on the feature distribution. Here, we show the results of more detailed analysis on the learned features. We used the same model as that used in Fig.~6 of the main paper.

\subsection{Fisher's criterion}
We calculated Fisher's criterion \cite{fisher1936use} for all combinations of two classes. Let $\{ {\bf x}_n\}_{n \in C_i}$ and ${\bf m}_i$ be features of class $C_i$ and the average of them ($\frac{1}{N_i} \Sigma_{n\in C_i}{{\bf x}_n}$), respectively. Here, Fisher's criterion between class $C_1$ and $C_2$ is defined as:
\begin{equation}
  \frac{{\bf w}^{\top} {\bf S}_B {\bf w}} {{\bf w}^{\top} {\bf S}_W {\bf w}},
  \vspace{-1mm}
\end{equation}
where:
\vspace{-1mm}
\begin{equation}
\begin{split}
  {\bf S}_B &= ({\bf m}_1 - {\bf m}_2)({\bf m}_1 - {\bf m}_2)^{\top}, \\
  {\bf S}_W &= \Sigma_{n \in C_1}{({\bf x}_n - {\bf m}_1)({\bf x}_n - {\bf m}_1)^{\top}} \\
  &+ \Sigma_{n \in C_2}{({\bf x}_n - {\bf m}_2)({\bf x}_n - {\bf m}_2)^{\top}}, \\
  {\bf w} &\propto {\bf S}_W^{-1}({\bf m}_1 - {\bf m}_2).
\end{split}
\vspace{-4mm}
\end{equation}

We show the mean Fisher's criterion in Table \ref{tab:fisher}. We used the activations of the $10$-th layer against training data. Fisher's criterion of the features learned with BC learning was larger than that of standard learning. This result shows that a discriminative feature space is learned with BC learning. 

\begin{table}[h]
	\centering
	\caption{Comparison of mean Fisher's criterion. BC learning indeed enlarges Fisher's criterion in the feature space.}
	\label{tab:fisher}
	\vspace{2mm}
	\small
	\begin{tabular}{lc}
		\toprule
		Learning & Mean Fisher's criterion \\
		\midrule
		Standard & $1.76$ \\
		BC (ours) & ${\bf 1.97}$ \\
		\bottomrule
	\end{tabular}
	\vspace{-2mm}
\end{table}

\subsection{Activation of final layer}
\begin{figure}
	\centering
	\includegraphics[width=\hsize, clip]{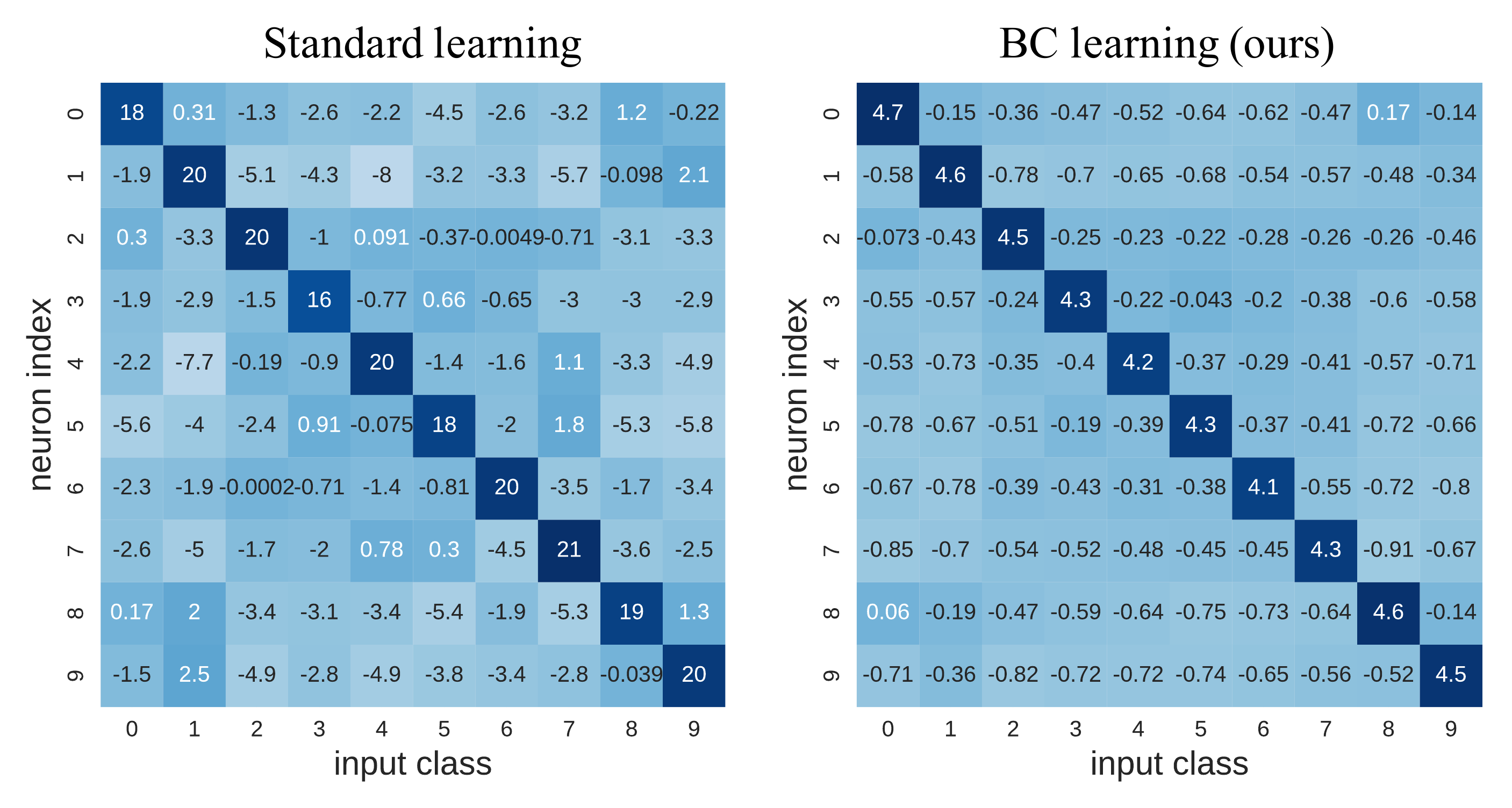}
	\vspace{-3mm}
	\caption{Activation of the final layer. The $(i,\,j)$ element represents the mean activation of $i$-th neuron of the final layer (before the softmax) against class $j$.}
	\label{fig:logit}
	\vspace{-2mm}
\end{figure}

We investigated the activation of the final layer against training images. The results are shown in Fig.~\ref{fig:logit}. The $(i,\,j)$ element represents the mean activation of $i$-th neuron of the final layer (before the softmax) against the training images of class $j$. As shown in this figure, each neuron responds only to the corresponding class when using BC learning. The responses against other classes are almost the same level and most of them are negative values, whereas the responses against classes other than the corresponding class have a large variance and some of them are positive values when using standard learning. This result indicates that the features of each class learned with BC learning are distributed in the opposite side of the features of other classes. Such features can be easily separated. It is possible that BC learning indeed regularizes the positional relationship among the feature distributions. 

\begin{figure}
	\centering
	\includegraphics[width=\hsize, clip]{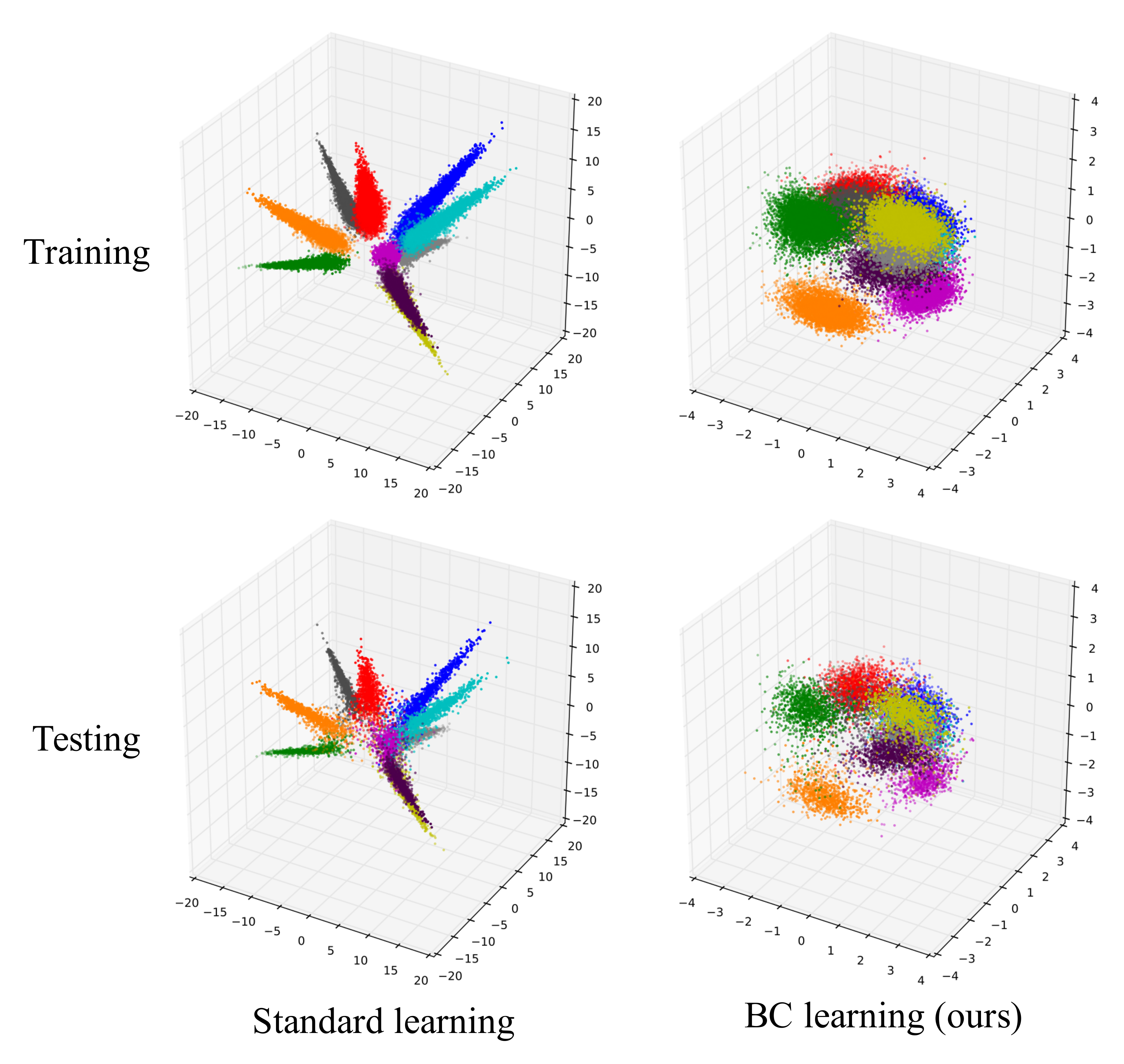}
	\vspace{-3mm}
	\caption{Feature distributions against training and testing data.}
	\label{fig:trainvstest}
	\vspace{-3mm}
\end{figure}

\begin{table*}
	\centering
	\caption{Learning settings for CIFAR experiments. The figures between brackets indicate the default learning settings.}
	\label{tab:settings}
	\vspace{2mm}
	\small
	\begin{tabular}{lccccc}
		\toprule
		Model & \# of epochs & Initial LR & LR schedule & nGPU & mini-batch size \\
		\midrule
		11-layer CNN & $250$ & $0.1$ & $\{ 100,\, 150,\, 200\}$ & $1$ & $128$ \\
		ResNet-29 \cite{xie2017aggregated} & ${\bf 375}$ ($300$) & ${\bf 0.0125}$ ($0.1$) & ${\bf{\{ 150,\, 225,\, 300\}}}$ ($\{ 150,\, 225\}$) & ${\bf 1}$ ($8$) & ${\bf 16}$ ($128$) \\
		ResNeXt-29 \cite{xie2017aggregated} & ${\bf 375}$ ($300$) & $0.1$ & ${\bf \{ 150,\, 225,\, 300\}}$ ($\{ 150,\, 225\}$) & $8$ & $128$ \\
		DenseNet \cite{huang2017densely} & ${\bf 375}$ ($300$) & $0.1$ & ${\bf \{ 150,\, 225,\, 300\}}$ ($\{ 150,\, 225\}$) & $4$ & $64$ \\
		Shake-Shake \cite{gastaldi2017shake} (CIFAR-10) & $1800$ & $0.2$ & cosine & $2$ & $128$ \\
		Shake-Shake \cite{gastaldi2017shake} (CIFAR-100) & $1800$ & $0.025$ & cosine & $2$ & $32$ \\
		\bottomrule
	\end{tabular}
\end{table*}

\subsection{Training features {\it vs.} testing features}

We compared the feature distributions against training and testing data. We visualized the activations of the $10$-th layer using $3$-D PCA. The results are shown in Fig.~\ref{fig:trainvstest}. The feature distributions of BC learning against training and testing data have similar shapes, whereas some testing examples are projected onto the points near the origin when using standard learning. This result indicates that the model learned with BC learning has a higher generalization ability.

\section{Details of CIFAR experiments}
\subsection{Learning settings}
We summarize the learning settings for CIFAR experiments in Table \ref{tab:settings}. Although most of them follow the original learning settings in \cite{xie2017aggregated, huang2017densely, gastaldi2017shake}, we slightly modified the learning settings for ResNet-29, ResNeXt-29 \cite{xie2017aggregated}, and DenseNet \cite{huang2017densely} in order to achieve a satisfactory performance. We trained the model by beginning
with a learning rate of {\it Initial LR}, and then divided the learning rate by $10$ at the epoch listed in {\it LR
schedule}, except that a cosine learning rate scheduling was used for Shake-Shake. We then terminated training after {\it \# of epochs} epochs.

\subsection{Configuration of 11-layer CNN}
We show the configuration of 11-layer CNN in Table \ref{tab:cnn}. We applied ReLU activation for all hidden layers and batch normalization \cite{ioffe2015batch} to the output of all convolutional layers. We also applied $0.5$ of dropout \cite{srivastava2014dropout} to the output of fc4 and fc5. We used a weight initialization of \cite{he2015delving} for all convolutional layers. We initialized the weights of each fully connected layer using the uniform distribution $U(-\sqrt{1/n},\, \sqrt{1/n}) $, where $n$ is the input dimension of the layer.
By using BC learning, we can achieve around $5.2\%$ and $23.7\%$ error rates on CIFAR-10 and CIFAR-100, respectively, despite the simple architecture.

\begin{table}[h]
	\centering
	\vspace{2mm}
	\caption{Configuration of 11-layer CNN.}
	\label{tab:cnn}
	\vspace{2mm}
	\small
	\begin{tabular}{lccccc}
		\toprule
		Layer  & ksize & stride & pad &\# filters & Data shape \\
		\midrule
		Input  & & & & & $(3,\ 32,\,32)$ \\
		\midrule
		conv1-1 & $3$ & $1$ &$1$ & $64$ & \\
		conv1-2 & $3$ & $1$ &$1$ & $64$ \\
		pool1  & $2$ & $2$ & & & $(64,\, 16,\, 16)$\\
		\midrule
		conv2-1 & $3$ & $1$ &$1$ & $128$ & \\
		conv2-2 & $3$ & $1$ &$1$ & $128$ \\
		pool2  & $2$ & $2$ & & & $(128,\, 8,\, 8)$\\
		\midrule
		conv3-1 & $3$ & $1$ &$1$ & $256$ & \\
		conv3-2 & $3$ & $1$ &$1$ & $256$ \\
		conv3-3 & $3$ & $1$ &$1$ & $256$ & \\
		conv3-4 & $3$ & $1$ &$1$ & $256$ \\
		pool3  & $2$ & $2$ & & & $(256,\, 4,\, 4)$\\
		\midrule
		fc4 & & & & $1024$ & $(1024,)$ \\
		fc5 & & & & $1024$ & $(1024,)$ \\
		fc6 & & & & \# classes & $($\# classes$,)$ \\
		\bottomrule
	\end{tabular}
\end{table}

\end{document}